%% file: MICCAI2026-main_conference_paper_template.tex
\documentclass[runningheads]{llncs}
\usepackage[T1]{fontenc}
\usepackage{booktabs}
\usepackage{amssymb}
\usepackage{amsmath}
\usepackage{graphicx,verbatim}
\usepackage{multirow}
\usepackage{mathtools}
\usepackage{hyperref}
\usepackage{cleveref}
\usepackage{siunitx} 
\usepackage{adjustbox}
\usepackage[table]{xcolor}
\usepackage{colortbl}
\usepackage{xcolor}
\newcommand{\etal}{\textit{et al.}}

\usepackage[most]{tcolorbox}

\definecolor{cardbg}{HTML}{EFF2F8}    
\definecolor{cardrule}{HTML}{C9D0DE}  
\definecolor{cardink}{HTML}{111318}   
\definecolor{cardmuted}{HTML}{4A5160} 
\definecolor{linkblue}{HTML}{2455D6}  

\newtcolorbox{titlecard}{%
  colback=cardbg, colframe=cardbg, boxrule=0pt,
  arc=5mm, outer arc=5mm,
  boxsep=0pt, left=7mm, right=7mm, top=5mm, bottom=5mm,
  width=\textwidth,
  before skip=0pt, after skip=4mm
}

\renewenvironment{abstract}{\par\small\noindent\ignorespaces}{\par}

\makeatletter
\def\@maketitle{\markboth{}{}%
  \setbox0=\vbox{\setcounter{@auth}{1}\def\and{\stepcounter{@auth}}%
                 \def\thanks##1{}\@author}%
  \global\value{@inst}=\value{@auth}%
  \global\value{auco}=\value{@auth}%
  \setcounter{@auth}{1}%
  \setbox0=\vbox{\institutename}}
\makeatother

\hypersetup{colorlinks=true, linkcolor=black, citecolor=black,
            filecolor=black, urlcolor=linkblue, breaklinks=true}

\usepackage{array}

\newcommand{\cardlink}[2]{%
  \textbf{#1:} & \href{#2}{\nolinkurl{#2}}\\[1.5pt]}

\usepackage{tikz}
\newlength{\cardlogowidth}
\newcommand{\cardlogo}{\includegraphics[height=12mm]{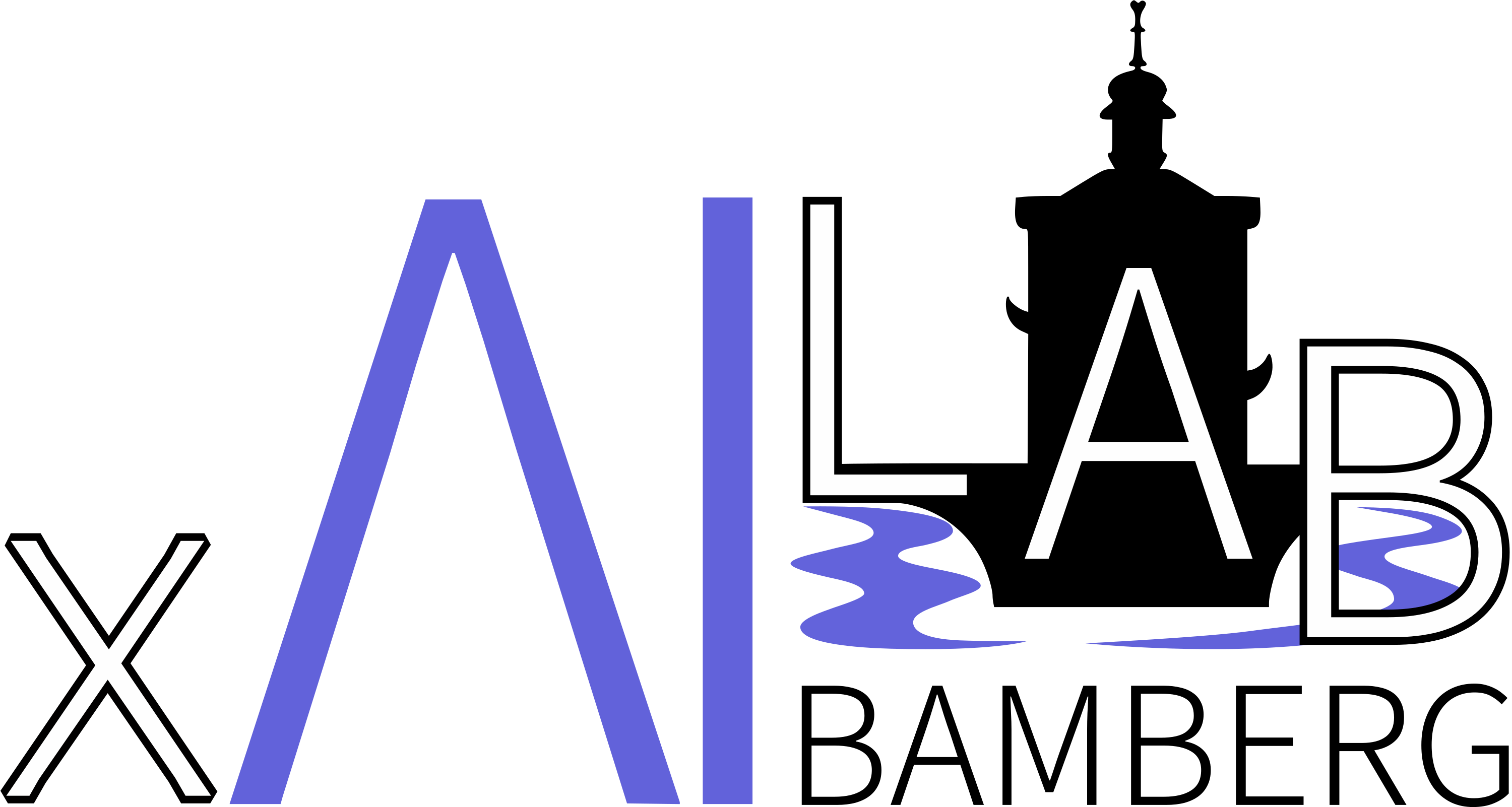}}
\usepackage[margin=1.25in]{geometry}
\begin{document}
\title{Layer Selection in VLMs for Zero-Shot OOD Detection via Multi-Resolution Entropy Estimation}
\titlerunning{Layer Selection in VLMs for Zero-Shot OOD Detection}
%

\author{Shyam Nandan Rai,  Francesco Di Salvo, Sebastian Doerrich, and
\\ Christian Ledig}  
\authorrunning{Rai et al.}
\institute{xAILab Bamberg, University of Bamberg, Bamberg, Germany \\
    \email{shyam.rai@uni-bamberg.de}}
  
\maketitle              
%


\begin{titlecard}
\color{cardink}
\setlength{\parindent}{0pt}

{\raggedright\hyphenpenalty=10000\exhyphenpenalty=10000
 \Large\bfseries\boldmath
 Layer Selection in VLMs for Zero-Shot OOD Detection via
 Multi-Resolution Entropy Estimation\par}

\vspace{4mm}

{\raggedright\normalsize\bfseries
 Shyam Nandan Rai, Francesco Di Salvo, Sebastian Doerrich, Christian Ledig\par}

\vspace{2.5mm}

{\raggedright\small xAILab Bamberg, University of Bamberg, Bamberg, Germany\par}
{\raggedright\footnotesize\color{cardmuted}\href{mailto:shyam.rai@uni-bamberg.de}{\nolinkurl{}}\par}

\vspace{3.5mm}
{\color{cardrule}\hrule height 0.5pt}
\vspace{3.5mm}

\begin{abstract}

Out-of-distribution (OOD) detection is crucial for safe deployment of medical AI systems, where domain shifts arise across institutions, acquisition protocols, and patient populations.
VLMs enable zero-shot OOD detection by embedding images into a language-aligned latent space, where cross-modal similarity serves as a non-parametric confidence signal for identifying in-distribution samples.
Yet existing methods rely almost exclusively on final-layer embeddings, implicitly assuming that the deepest representations are universally optimal. We first show that this assumption does not hold in medical imaging: intermediate layers provide complementary OOD signals, and the optimal representational depth depends on the respective image modality. While prior work selects layer combinations via entropy minimization of normalized histograms, we demonstrate that single-resolution entropy estimation is highly sensitive to binning choices, leading to performance variations of up to $19.3\%$ AUROC. To address this instability, we propose a multi-resolution entropy estimation strategy that aggregates histogram statistics across multiple discretization scales, enabling robust and stable intermediate-layer selection. Across two medical OOD benchmarks, namely MIDOG and OASIS, covering distinct imaging modalities, diverse shift types, and different VLM backbones, our method consistently outperforms state-of-the-art approaches, offering a lightweight and stable solution for zero-shot OOD detection.


\end{abstract}

\vspace{3.5mm}
{\color{cardrule}\hrule height 0.5pt}
\vspace{3.5mm}

\noindent
\begin{minipage}[b]{\dimexpr\linewidth-\cardlogowidth-3mm\relax}
{\small
\begin{tabular}[b]{@{}l@{\hspace{1.2em}}>{\raggedright\arraybackslash}p{\dimexpr\linewidth-5.2em\relax}@{}}
\cardlink{Website}{https://shyam671.github.io/layer-selection-ood/}
\cardlink{Contact}{shyam.rai@uni-bamberg.de}
\end{tabular}\par}
\end{minipage}\hfill
\begin{minipage}[b]{\cardlogowidth}
\raggedleft\cardlogo
\end{minipage}

\end{titlecard}


\section{Introduction}

Out-of-distribution (OOD) detection~\cite{yang2024generalized} is a critical research problem in healthcare, where models are deployed across diverse institutions, imaging devices, acquisition protocols, and heterogeneous patient populations~\cite{stacke2020measuring,finlayson2021clinician,hong2024out}. Vision-Language Models (VLMs)~\cite{CLIP,biomedclip,khattak2024unimed} enable zero-shot OOD detection by measuring the alignment between image and text embeddings. This is particularly beneficial because the in-distribution manifold can be characterized through semantically rich textual descriptions, while cross-modal similarity serves as a proxy for prediction confidence. However, existing zero-shot OOD detection methods based on VLMs~\cite{MCM,JuLie_Delving_MICCAI2025} primarily rely on final-layer representations, implicitly assuming that they are universally optimal for OOD detection. This assumption has recently been challenged for VLMs~\cite{de2025mysteries}, as well as in other domains~\cite{wei2025xmahalanobis,jelenic2024outofdistribution,darrin2024unsupervised,lambert2023multi}. 
De la Jara~\etal~\cite{de2025mysteries} show for the natural image domain that OOD-discriminative signals are distributed across multiple representational depths within VLMs, rather than being confined to the final layer. In particular, intermediate layers have been shown to capture diverse visual features~\cite{raghu2021do} that complement high-level semantic representations. Despite these findings, a systematic investigation into whether aggregating layer-wise OOD scores can substantially improve detection performance in the context of medical image analysis remains unexplored. We therefore revisit the problem of zero-shot medical OOD detection and ask: \textit{Can intermediate layer representations be systematically utilized to enhance zero-shot OOD detection across modalities?} 

\begin{figure}
	\centering
	\includegraphics[width=1\linewidth]{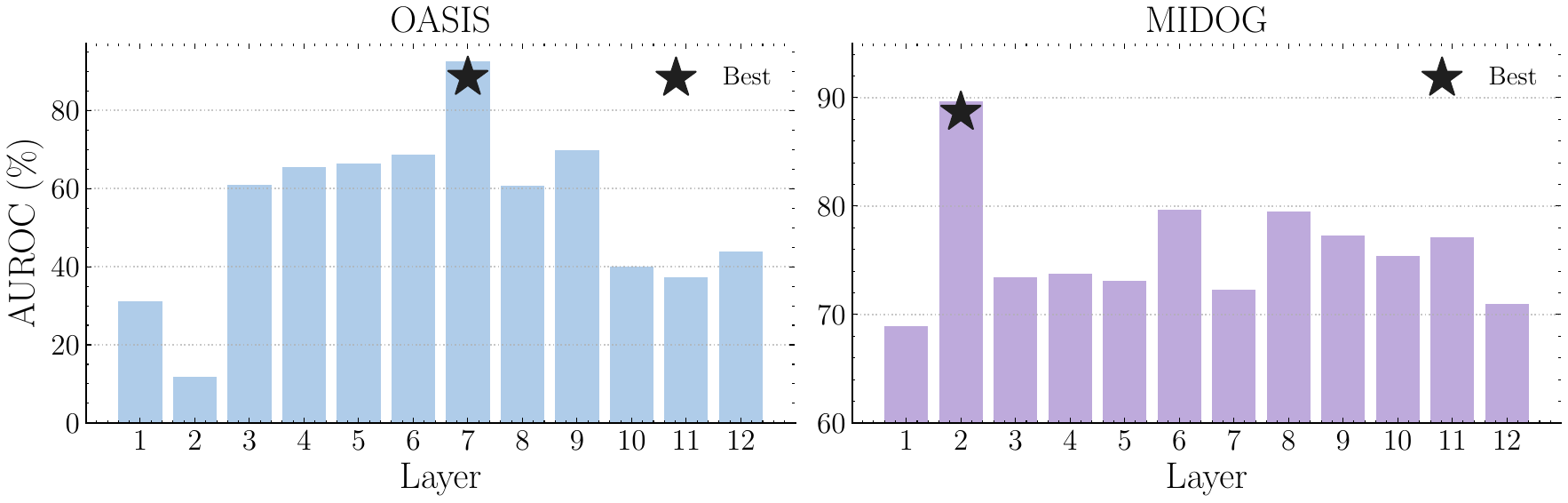}
	\caption{Layer-wise zero-shot OOD detection performance (AUROC, \%), using UniMedCLIP. \textbf{Left}: OASIS (brain MRI). \textbf{Right}: MIDOG (histopathology). Each bar corresponds to the OOD detection performance using Maximum Concept Matching (MCM) for the corresponding encoder layer, averaging for both \textit{near}- and \textit{far}-OOD shifts. The results demonstrate that the best layer depends on the dataset, \textit{i.e.}, imaging modality. Notably, the final layer does not exhibit the best performance in either case.}
    \label{fig:intermediate_layers}
\end{figure}

To answer this question, we perform a systematic layer-wise analysis of zero-shot OOD detection across multiple medical imaging datasets, presented in ~\Cref{fig:intermediate_layers}. Our results reveal that the optimal representational depth is modality-dependent: texture-driven shifts in histopathology \cite{kong2011partitioning} are primarily captured in earlier layers, whereas semantic and anatomical shifts in brain MRI~\cite{desgranges2007anatomical} benefit from the global contextual representations encoded in deeper layers.

Motivated by these observations, we revisit entropy-based intermediate-layer selection~\cite{de2025mysteries} in the context of medical imaging. While originally developed for large-scale natural image benchmarks, we find that its reliance on single-resolution histogram entropy leads to unstable performance in small-scale medical datasets, where discretization sensitivity can degrade AUROC by up to $19.3\%$ (\textit{cf}. ~\Cref{fig:ab-bins}).
To address this limitation, we propose a multi-resolution entropy estimation strategy that aggregates entropy estimates across multiple discretization scales of histogram, enabling robust and stable layer selection. We evaluate our method on two medical OOD benchmarks using two VLM backbones, covering distinct imaging modalities and varying degrees of distribution shift. Our approach substantially reduces bin-size sensitivity and consistently improves OOD detection performance over state-of-the-art zero-shot baselines, demonstrating strong generalization across backbones and modalities. In summary:
\begin{itemize}
    \item To the best of our knowledge, we provide the first systematic layer-wise analysis of VLMs for zero-shot OOD detection in medical image analysis, revealing that OOD-discriminative signals emerge at modality-dependent representational depths.
    \item We identify the instability of single-resolution entropy-based layer selection in data-scarce medical settings and introduce a multi-resolution entropy estimation strategy that yields robust layer selection.
    \item Extensive experiments across two medical OOD benchmarks and two VLM backbones demonstrate consistent performance improvements over state-of-the-art zero-shot baselines, spanning multiple imaging modalities and distribution shifts.
\end{itemize}

\section{Intermediate layers for medical OOD detection} 

In this section, we investigate whether intermediate representations provide complementary signals for zero-shot OOD detection. We use UniMedCLIP~\cite{khattak2024unimed} as the reference backbone. Let $I$ denote an input image. The visual encoder $E$ consists of $N$ sequential layers, producing intermediate representations $\{L_j(I)\}_{j=1}^N$, where $L_j(I)$ denotes the output of the $j$-th layer and $L_N(I)=E(I)$ corresponds to the canonical final-layer embedding. To enable layer-wise comparison within the shared image–text embedding space, each intermediate representation is projected using the same pretrained projection head as the final layer~\cite{de2025mysteries}. We used the datasets as described in Section~\ref{Experimentation} and Maximum Concept Matching (MCM)~\cite{MCM} as OOD scoring function.

Figure~\ref{fig:intermediate_layers} shows that the final-layer embedding does not exhibit optimal performance. For MIDOG (histopathology), the strongest results are obtained in early layers. This indicates that fine-grained, texture-dominant features are particularly informative for detecting distribution shifts. In contrast, for OASIS (brain MRI), characterized by structured anatomical patterns, mid-level layers provide better performance. These observations demonstrate that OOD signals emerge at different representational depths depending on modality, motivating intermediate-layer selection rather than reliance on the final-layer representation.

\begin{figure}
	\centering
    \includegraphics[width=1\linewidth]{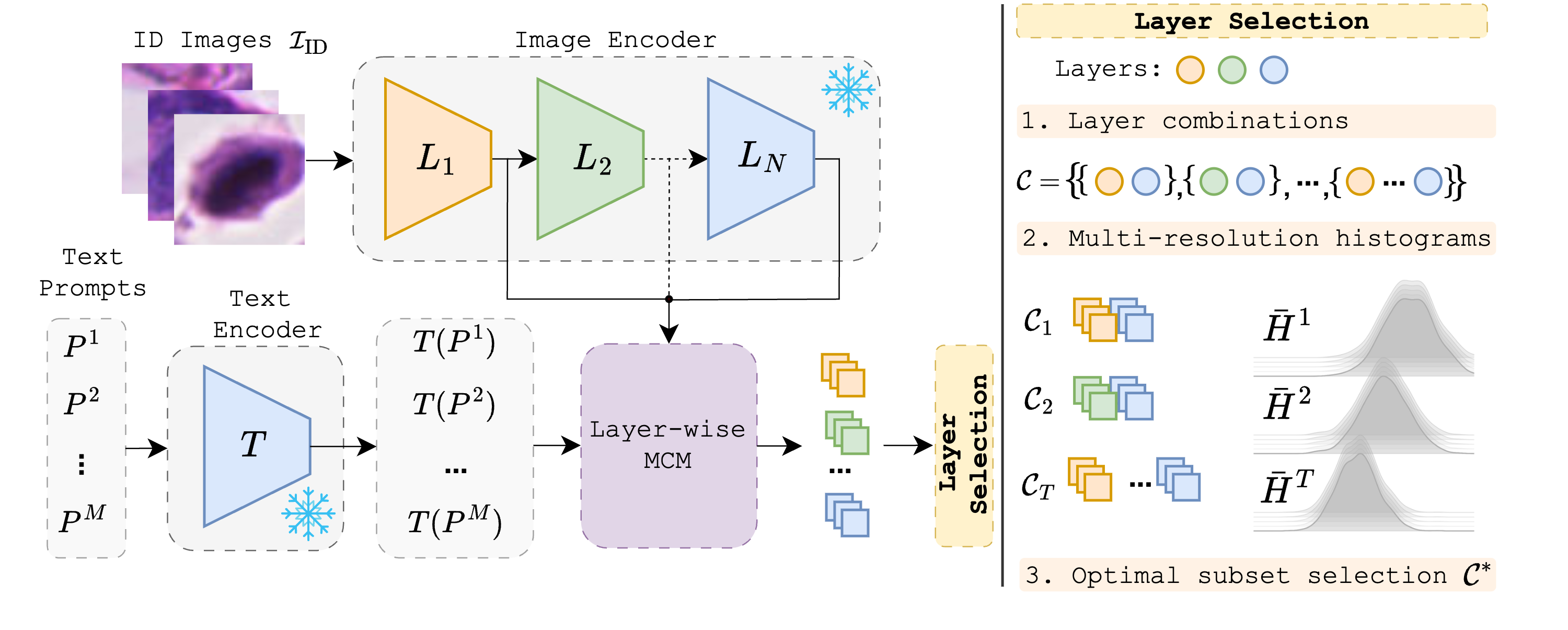}
	\caption{Overview of the proposed intermediate-layer selection framework. Given in-distribution images $I \in \mathcal{I}_{\mathrm{ID}}$ and prompts $\{P^1,\dots,P^M\}$, the visual encoder outputs intermediate representations $\{L_j(I)\}_{j=1}^N$, which are projected into the shared image–text embedding space and scored via layer-wise MCM: $S_{\mathrm{MCM}}^{(j)}(I)$. Candidate layer subsets $\mathcal{C}$, always including the final layer $N$, are constructed, and their aggregated scores are collected across the in-distribution dataset to form the subset score distribution $S_{\mathcal{C}}$. This distribution is evaluated via multi-resolution entropy $\bar{H}(S_{\mathcal{C}})$, and the optimal subset $\mathcal{C}^*$ minimizing aggregated entropy is selected for test-time inference.}
    \label{fig:pull_figure}
\end{figure}

\section{Method}

Motivated by our observation that OOD signals emerge at modality-dependent representational depths (\Cref{fig:intermediate_layers}), we propose a robust intermediate-layer selection method based on multi-resolution entropy estimation. 

\subsection{Preliminaries}

VLMs learn a shared embedding space for images and text through contrastive pretraining. We consider CLIP-style models~\cite{CLIP} consisting of an image encoder $E: \mathcal{I} \rightarrow \mathbb{R}^d$ and a text encoder $T: \mathcal{P} \rightarrow \mathbb{R}^d$. Given a set of prompts describing the in-distribution concepts $\{P^1, \dots, P^M\}$, the similarity between an input image $I$ and prompt $P^j$ is computed via cosine similarity in the shared embedding space. The corresponding softmax probability is:

\begin{equation}
p(y^j \mid I) =
\frac{
\exp\left( \cos(E(I), T(P^j)) / \tau \right)
}{
\sum_{m=1}^{M}
\exp\left( \cos(E(I), T(P^m)) / \tau \right)
}.
\end{equation}

Maximum Concept Matching (MCM)~\cite{MCM} defines the OOD scoring function as: $S_{\text{MCM}}(I) = \max_j \, p(y^j \mid I)$
which is equivalent to the Maximum Softmax Probability (MSP)~\cite{hendrycks2017a} used in vision models, but computed over image–text similarities. An input is in in-distribution if $S_{\text{MCM}}(I) \geq \theta$, and OOD otherwise. 

\subsection{Multi-resolution entropy estimation}

Let $\mathcal{I}_{\mathrm{ID}}$ denote the in-distribution image set and 
$\{P^1,\dots,P^M\}$ the corresponding prompts, where $M$ denotes the number of available prompts. 
For each image $I \in \mathcal{I}_{\mathrm{ID}}$, the visual encoder produces intermediate representations $\{L_j(I)\}_{j=1}^N$, which are projected into the shared image–text embedding space. $N$ denotes the total number of representations from an image. For each layer $j$, we compute the layer-wise MCM represented as $S_{\mathrm{MCM}}^{(j)}(I)$. To aggregate complementary signals across representational depths, we evaluate all possible layer subsets while enforcing inclusion of the final layer~\cite{de2025mysteries}. Each candidate subset is defined as
$\mathcal{C} = \{N\} \cup \mathcal{C}'$, where $\mathcal{C}' \subseteq \{1,\dots,N-1\}$. For a given subset $\mathcal{C}$, we form the score distribution used for entropy estimation by collecting the aggregated scores across the in-distribution images as:

\begin{equation}
S_{\mathcal{C}} = \left\{
\frac{1}{|\mathcal{C}|}
\sum_{j \in \mathcal{C}}
S_{\mathrm{MCM}}^{(j)}(I) \right\}_{\forall I \in \mathcal{I}_{\mathrm{ID}}}.
\end{equation}
\noindent\textbf{Ensemble of histograms} \, To obtain a robust uncertainty measure, we compute entropy across an ensemble of histograms $\{{B_k\}}_{k=1}^K$, where $K$ denotes the number of binning scales. Formally, entropy for a histogram is calculated as:

\begin{equation}
H_k(S_{\mathcal{C}})
=
- \sum_{b=1}^{B_k}
p_b^{(k)} \log p_b^{(k)}.
\end{equation}
The entropy estimates are then averaged across histogram resolutions,
\begin{equation}
\bar{H}(S_{\mathcal{C}})
=
\frac{1}{K}
\sum_{k=1}^{K}
H_k(S_{\mathcal{C}}),
\end{equation}

We select the optimal layer subset as $\mathcal{C}^*= \arg\min_{\mathcal{C}} \bar{H}(S_{\mathcal{C}})$. Ablation studies in Section \ref{sec:ablation_sensitivity} demonstrate that multi-resolution entropy aggregation substantially reduces bin-size sensitivity compared to single-resolution estimation. \newline

\noindent\textbf{Test-time inference} \,
During inference, we use $\mathcal{C}^*$ along with the final layer to compute MCM scores. $\mathcal{C}^*$ is determined once on the ID set and remains the same for every test sample.

\section{Experimentation}
\label{Experimentation}
\subsubsection{Baselines} 
We compare our method against MCM~\cite{MCM} and Ju \etal~\cite{JuLie_Delving_MICCAI2025}, which perform zero-shot OOD detection using final-layer embeddings. Ju \etal\ improve performance by hierarchical prompting. We also include De la Jara \etal~\cite{de2025mysteries}, the closest related approach, which selects intermediate layers via entropy minimization using a single histogram resolution, leading to discretization instability. 

\subsubsection{Datasets and metrics} 
For evaluation, we adapt the recently proposed OpenMIBOOD benchmark~\cite{gutbrod2025openmibood} for VLM backbones, focusing on histopathology and brain MRI as representative medical imaging modalities. We excluded PhaKIR~\cite{rueckert2026comparative} dataset, as none of the VLM backbones were pretrained on surgical data, making predictions on this dataset unreliable. Our analysis centers on near- and far-OOD distribution shifts, enabling evaluation across varying degrees of semantic distribution shift within modality. For histopathology, we use MIDOG~\cite{MIDOG} (CC BY 4.0), where the in-distribution data consist of mitotic and non-mitotic cell crops extracted from H\&E-stained whole-slide images. Near-OOD datasets introduce semantic shifts across cancer types, species, and acquisition settings, while far-OOD samples originate from 
cervical cancer cell images \cite{CCAgT} (CCAgT, Apache 2.0) and breast FNAC cytology \cite{FNAC} (license not disclosed). For brain MRI, OASIS3-MRI \cite{lamontagne2019oasis} (custom license) serves as the in-distribution dataset, with OASIS3-CT used as near-OOD and MSD-H~\cite{MSD-h} (BSD 2-Clause) and CHAOS~\cite{chaos} (CC BY-NC-SA 4.0) as far-OOD datasets. OOD detection performance is evaluated using the Area Under the Receiver Operating Characteristic curve (AUROC) and the False Positive Rate at $95\%$ True Positive Rate (FPR95).

\subsubsection{Implementation details} We use two CLIP-based medical VLMs: BioMedCLIP~\cite{biomedclip} and UniMedCLIP~\cite{khattak2024unimed}. Each of them used ViT-B/16 backbone as the vision model. We perform all the evaluations in a training-free and inference-only setting. The temperature parameter $\tau$ is fixed to $1.0$ across all evaluations. In-distribution prompts are generated using a large-scale language model~\cite{achiam2023gpt}, following prior work~\cite{JuLie_Delving_MICCAI2025}. Unlike Ju \etal~\cite{JuLie_Delving_MICCAI2025}, which employ hierarchical prompt generation, we adopt a flat prompt structure to reduce redundancy. For fair comparison, the same prompt set is used for our method and all remaining baselines. We employ $9$ prompts for MIDOG and $4$ prompts for OASIS. An example prompt includes `a healthy brain MRI' for OASIS and `a diagnostic histopathology slide showing breast carcinoma' for MIDOG. For De la Jara \etal~\cite{de2025mysteries}, we use 16 histogram bins, as recommended by the authors. Our multi-resolution entropy estimation utilizes a histogram ensemble with bin sizes $\{4, 8, 16, 32, 64\}$. For our method and De la Jara \etal~\cite{de2025mysteries}, we constrain the layer subset size to at most four layers (including the final layer) and select the subset that minimizes the respective entropy criterion using only in-distribution data.

\input{main_tab_v2}

\subsubsection{Results}
Table~\ref{tab:ood_exact} summarizes OOD detection performance across UniMedCLIP and BioMedCLIP backbones on the MIDOG and OASIS benchmarks. On OASIS, with both backbones, our method achieves near-perfect performance across both near- and far-OOD settings and. Notably, the largest performance gains are observed in FPR95: on BioMedCLIP, our method reduces FPR95 by approximately $20\%$ in both near- and far-OOD compared to the strongest baseline. On the MIDOG benchmark, our method demonstrates the best overall results with the UniMedCLIP backbone, achieving the best near-OOD performance (AUROC: $54.6\%$, FPR95: $94.4\%$) and far-OOD performance (AUROC: $97.4\%$, FPR95: $13.4\%$). With BioMedCLIP, results are more mixed across methods. For instance, MCM \cite{MCM} achieves the best results on near-OOD, while De la Jara~\etal\cite{de2025mysteries} achieve the best far-OOD results, suggesting that no single method dominates consistently, making comparisons on this backbone less conclusive. Overall, UniMedCLIP emerges as the more reliable backbone for OOD detection in our evaluation, yielding stronger and more stable results across both benchmarks. 
Within this setting, our method achieves the best or competitive performance in nearly all configurations, highlighting its effectiveness and robustness for medical OOD detection.


\section{Ablation studies}

We ablate two fundamental components of our method: (1) the impact of the number of aggregated intermediate layers, and (2) the robustness of multi-resolution entropy estimation compared to single-resolution histogram binning. All experiments are performed with UniMedCLIP unless stated otherwise.

\subsection{Number of selected layers}

\begin{figure}
	\centering
    \includegraphics[width=0.9\linewidth]{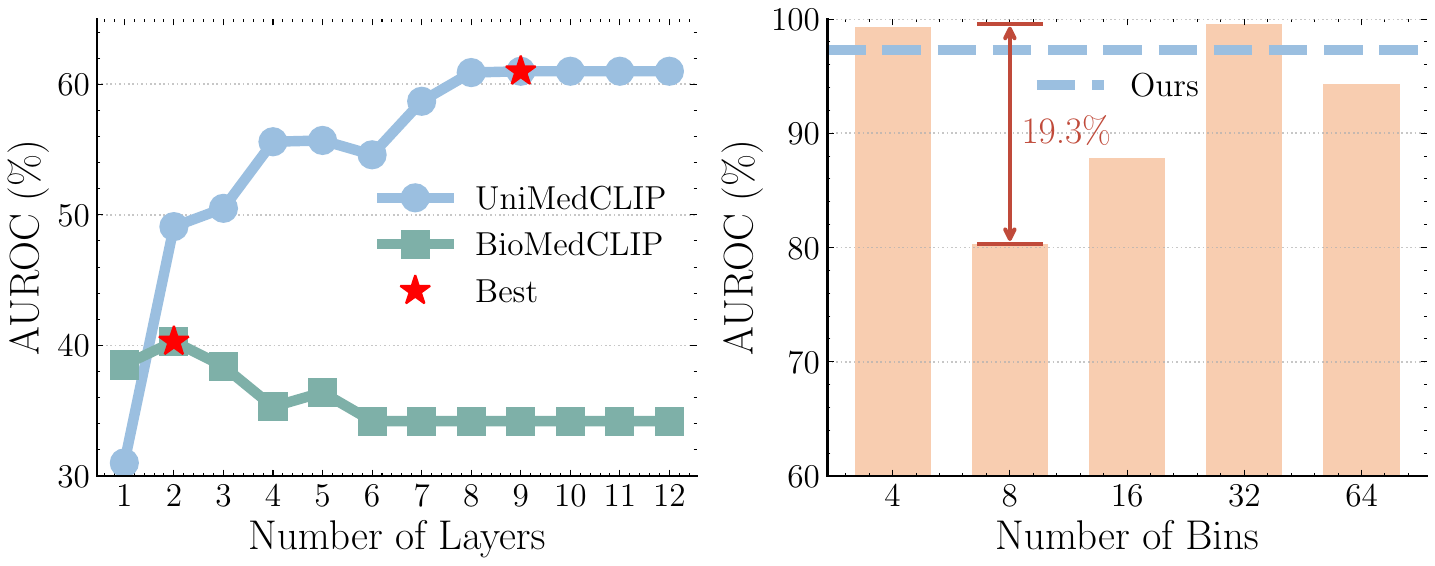}
	\caption{\label{fig:ablations}\textbf{Left}: OOD performance on MIDOG (near) as a function of the number of aggregated intermediate layers. \textbf{Right}: Sensitivity analysis on OASIS (near) of single-resolution entropy-based layer selection across different histogram bin sizes, where performance varies by up to $19.3\%$. In contrast, our multi-resolution entropy estimation strategy achieves stable and near-optimal performance.}
    \label{fig:ab-bins}
\end{figure}

We study how performance varies as additional intermediate layers are incorporated into the aggregation. Starting from the final layer alone, we progressively add layers selected by our entropy criterion. We report the analysis on MIDOG (near), because OASIS produces near-saturated AUROC in this setting, leaving little headroom to measure the marginal gains from adding intermediate layers. Figure~\ref{fig:ablations} (left) shows that aggregating more intermediate representations consistently improves OOD detection for UniMedCLIP, increasing AUROC above the last-layer baseline. However, in BioMedCLIP, AUROC eventually decreases as more intermediate layers are aggregated, showing layer selection is beneficial compared to using all layers. This confirms that OOD-relevant signals are distributed across layers rather than confined to the final layer.


%
\subsection{Sensitivity of histogram bins}
\label{sec:ablation_sensitivity}
We next evaluate the sensitivity of entropy-based selection to the histogram bin count $B$. Following De la Jara \etal~\cite{de2025mysteries}, we implement a single-resolution entropy baseline and sweep recommended bin counts $\{4,8,16,32,64\}$. We conduct the analysis on OASIS (near), reflecting a realistic medical setting with limited in-distribution data (${<}1{,}000$ samples). As shown in Figure ~\ref{fig:ablations} (right), OOD performance varies substantially across bin sizes (standard deviation $\sigma=8.2\%$) with a maximum drop of $19.3\%$, highlighting the fragility of single-resolution entropy estimation. In contrast, our multi-resolution ensemble-of-histograms achieves stable, near-optimal performance without requiring manual bin-size selection.

\section{Discussion}

\noindent\textbf{Limitations} \, Our approach remains dependent on the underlying VLM backbone, and although we evaluate two widely established and important medical modalities, further validation across additional imaging domains is required. Moreover, we employ uniform averaging for layer aggregation, which may not capture potentially optimal weighted combinations. Future works could explore finer-grained scoring strategies beyond MCM, for instance, utilizing the entire prompt-wise score distributions. \newline

\noindent\textbf{Conclusions} \, In this work, we demonstrate that OOD signals in medical Vision–Language Models emerge at modality-dependent representational depths, challenging the common reliance on final-layer embeddings. Through a systematic layer-wise analysis, we reveal substantial instability in single-resolution entropy-based layer selection and show that binning choices can significantly affect performance. To address this, we introduce a multi-resolution entropy estimation strategy that enables robust intermediate-layer aggregation. Across datasets and backbones, our results show that our method offers a simple yet effective strategy for enhancing zero-shot OOD detection.\newline

\noindent\textbf{Acknowledgments} The authors gratefully acknowledge the scientific support and HPC resources provided by the Erlangen National High Performance Computing Center (NHR@FAU) of the Friedrich-Alexander-Universität Erlangen-Nürnberg (FAU). The hardware is funded by the German Research Foundation (DFG). This study was further funded through the Hightech Agenda Bayern (HTA) of the Free State of Bavaria, Germany.\newline

\noindent\textbf{Disclosure of Interests} The authors have no competing interests to declare that are relevant to the content of this article.
%
%
%
\bibliographystyle{splncs04}
\bibliography{mybibliography}
%






\end{document}

%% file: main_tab_v2.tex
\begin{table}[ht]
\centering
\small
\setlength{\tabcolsep}{1pt} 
\caption{OOD detection results (AUROC $\uparrow$, FPR95 $\downarrow$) across two backbones (UniMedCLIP and BioMedCLIP) and two datasets (MIDOG and OASIS), each evaluated under \textit{near}- and \textit{far}-OOD settings. Despite some variability on BioMedCLIP for MIDOG, our method consistently achieves strong performance and substantially outperforms reference approaches across most configurations.}
\label{tab:ood_exact}
\begin{adjustbox}{max width=\textwidth}
\begin{tabular}{lccccccccccccccccccc}
\toprule

    & \multicolumn{4}{c}{\cellcolor{gray!15} \texttt{UniMedCLIP}}
    & 
    & \multicolumn{4}{c}{\cellcolor{gray!15} \texttt{BioMedCLIP}}
\\ 


    & \multicolumn{2}{c}{\emph{Near-OOD}}
    & \multicolumn{2}{c}{\emph{Far-OOD}}
    & 
    & \multicolumn{2}{c}{\emph{Near-OOD}}
    & \multicolumn{2}{c}{\emph{Far-OOD}}
\\

\cmidrule(lr){2-3}
\cmidrule(lr){4-5}
\cmidrule(lr){7-8}
\cmidrule(lr){9-10}


    & AUROC $\uparrow$ & FPR95 $\downarrow$
    & AUROC $\uparrow$ & FPR95 $\downarrow$
    & 
    & AUROC $\uparrow$ & FPR95 $\downarrow$
    & AUROC $\uparrow$ & FPR95 $\downarrow$
    
\\

\midrule

\multicolumn{10}{c}{\cellcolor{gray!15} \texttt{\texttt{MIDOG}}}
\\
MCM~\cite{MCM} 
    & 50.5 & 94.5 & 84.3 & 49.3 
    & 
    & \textbf{52.7} & \textbf{95.6} & 46.3 & 99.1 \\

Ju~\etal~\cite{JuLie_Delving_MICCAI2025} 
    & 40.5 & 98.0 & 37.4 & 99.7 
    & 
    & 50.2 & 97.4 & 84.8 & 66.0 \\

De la Jara~\etal~\cite{de2025mysteries}  
    & 29.1 & 99.7 & 97.1 & 16.6
    & 
    & 32.1 & 97.3 & \textbf{99.8} & \textbf{1.3} \\

\textit{Ours}
    & \textbf{54.6} & \textbf{94.4} & \textbf{97.4} & \textbf{13.4} 
    & 
    & {36.4} & {100.0} & {99.5} & {2.7} \\ 

\midrule

\multicolumn{10}{c}{\cellcolor{gray!15} \texttt{\texttt{OASIS}}}
\\
MCM~\cite{MCM} 
    & 55.2& 99.9 & 8.6 & 100.0
    & 
    & 90.4 & 34.8 & 49.0 & 99.9 \\

Ju~\etal~\cite{JuLie_Delving_MICCAI2025}  
    & 25.3 & 100.0 & 27.0 & 99.2 
    & 
    & 85.7 & 34.8 & 92.9 & 20.4 \\

De la Jara~\etal~\cite{de2025mysteries} 
    & \textbf{100.0} & \textbf{0.0} & 97.7 & 6.5
    & 
    & 50.0 & 78.5 & 60.7 & 39.3 \\

\textit{Ours} 
    & \textbf{100.0} & \textbf{0.0} & \textbf{99.2} & \textbf{1.0} 
    & 
    & \textbf{97.3} & \textbf{14.3} & \textbf{100.0} & \textbf{0.0} \\ 

\bottomrule
\end{tabular}
\end{adjustbox}

\end{table}